\documentclass[10pt,twocolumn,letterpaper]{article}
\pdfoutput=1
\usepackage{cvpr}
\usepackage{times}
\usepackage{epsfig}
\usepackage{graphicx}
\usepackage{amsmath}
\usepackage{amssymb}
\usepackage{multirow}
\usepackage{booktabs}
\usepackage[accsupp]{axessibility} 

\usepackage[pagebackref=true,breaklinks=true,letterpaper=true,colorlinks,bookmarks=false]{hyperref}

\begin{document}

\makeatletter
\renewcommand*{\@fnsymbol}[1]{\ensuremath{\ifcase#1\or \dag\or \dagger\or \ddagger\or
   \mathsection\or \mathparagraph\or \|\or **\or \dagger\dagger
   \or \ddagger\ddagger \else\@ctrerr\fi}}
\makeatother

\title{MRA-GNN: Minutiae Relation-Aware Model over Graph Neural Network for Fingerprint Embedding}
\vspace{-1em}

\author{Yapeng Su,\ \ Tong Zhao\thanks{Corresponding author},\ \ Zicheng Zhang\\
School of Mathematical Sciences, University of Chinese Academy of Sciences, China\\
{\tt\small \{suyapeng21@mails., zhaotong@, zhangzicheng19@mails.\}ucas.ac.cn}}
\vspace{-1em}

\maketitle
\thispagestyle{empty}

\begin{abstract}
\vspace{-0.4em}
Deep learning has achieved remarkable results in fingerprint embedding, which plays a critical role in modern Automated Fingerprint Identification Systems. However, previous works including CNN-based and Transformer-based approaches fail to exploit the nonstructural data, such as topology and correlation in fingerprints, which is essential to facilitate the identifiability and robustness of embedding. To address this challenge, we propose a novel paradigm for fingerprint embedding, called Minutiae Relation-Aware model over Graph Neural Network (MRA-GNN). Our proposed approach incorporates a GNN-based framework in fingerprint embedding to encode the topology and correlation of fingerprints into descriptive features, achieving fingerprint representation in the form of graph embedding. Specifically, we reinterpret fingerprint data and their relative connections as vertices and edges respectively, and introduce a minutia graph and fingerprint graph to represent the topological relations and correlation structures of fingerprints. We equip MRA-GNN with a Topological relation Reasoning Module (TRM) and Correlation-Aware Module (CAM) to learn the fingerprint embedding from these graphs successfully. To tackle the over-smoothing problem in GNN models, we incorporate Feed-Forward Module and graph residual connections into proposed modules. The experimental results demonstrate that our proposed approach outperforms state-of-the-art methods on various fingerprint datasets, indicating the effectiveness of our approach in exploiting nonstructural information of fingerprints.
\vspace{-0.5em}
\end{abstract}


\begin{figure} [t]
	\centering
	{
		\includegraphics[width=1\linewidth]{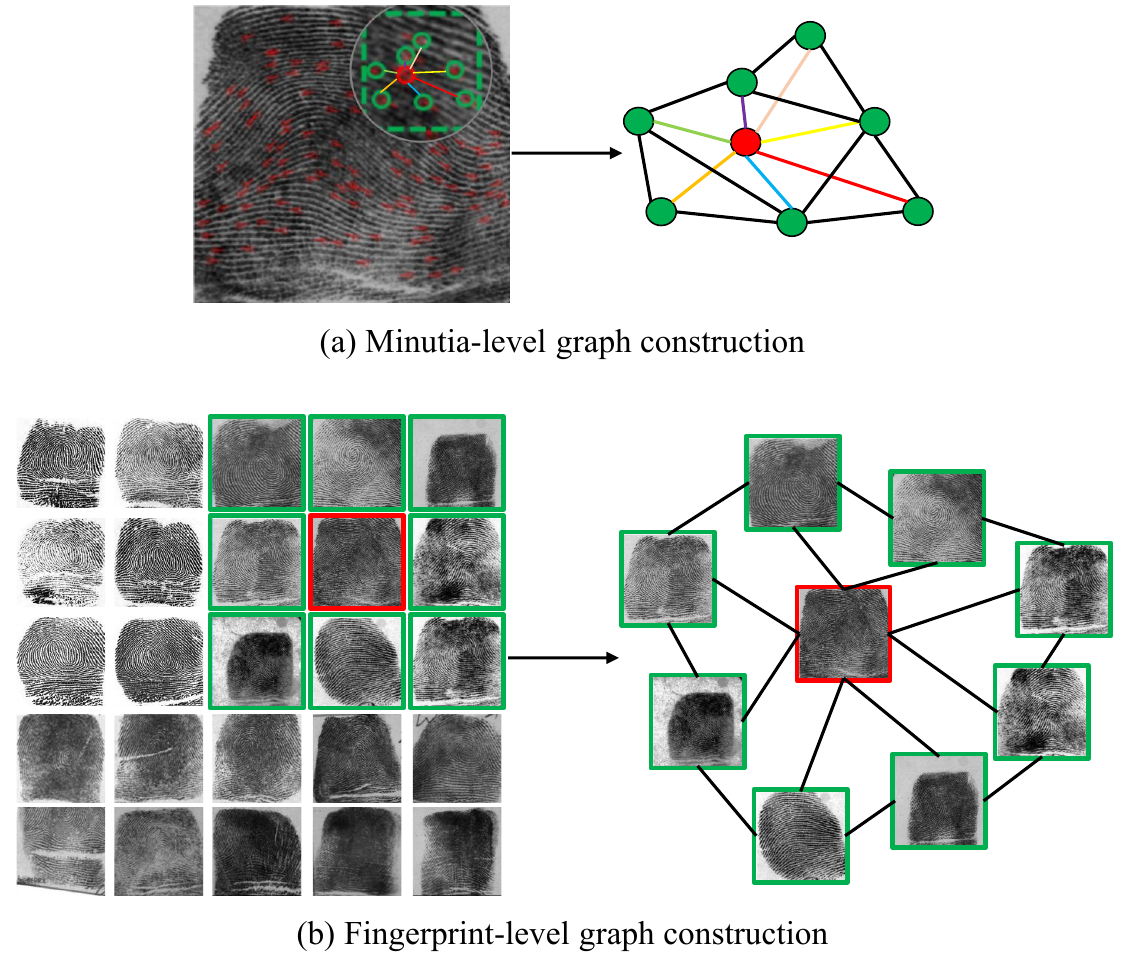}
	}
    \vspace{-0.7em}
	\caption{Illustration of graph construction in the proposed MRA-GNN. (a) Minutia-level graph construction: for each fingerprint, the minutiae and their neighborhood relations are considered to form vertices and edges of a graph, expressing the topology of a fingerprint. (b) Fingerprint-level graph construction: similarly, a batch of fingerprints is sampled to form a graph, expressing local structure in the fingerprint manifold. We show the central and nearby vertices in red and green color. The fingerprint embedding will be extracted from these graphs via MRA-GNN.}
	\label{fig:graph construct}
\vspace{-1.5em}
\end{figure}

\vspace{-1em}
\section{Introduction}
Thanks to the rapid development of biometrics \cite{triplets-1,LatentAFIS,Deepface,tai2023multi}, Automated Fingerprint Identification Systems (AFIS) have been applied in a variety of fields, such as civil identification and criminal identification. Fingerprint embedding, which refers to the encoding of fingerprints as compact and identifiable features, has a significant impact on the efficiency and accuracy of AFIS. Classical embedding algorithms generally focus on the geometric structures of fingerprint minutiae, \textit{e.g.}, triplets \cite{triplets-1,triplets-2} and cylinders \cite{cylinder-1}. Nevertheless, these methods heavily rely on handcrafted rules to extract embedded features, thereby suffering from inferior robustness and limited generalization in large-scale automated fingerprint recognition scenarios.

In recent years, deep learning methods have been introduced into AFIS and made significant advances in fixed-length fingerprint embedding beyond classical methods. Such deep learning methods can be primarily classified into two categories according to network structure, namely \textit{CNN-based} and \textit{Transformer-based} approaches. CNN-based approaches \cite{MDC,inceptionv3,Fingerpatches,Deepprint,LatentAFIS} generally involve multiple stages to extract feature representation from fingerprint images, such as global alignment, minutia detection, and patch extraction. This leads to a tedious training process that is difficult to extract global information from given fingerprint images. Inspired by the success of Vision Transformer, recent Transformer-based works~\cite{transformer-based1, transformer-based2} show that the Transformer with powerful feature extraction capability can learn better fingerprint embedding with global representation in an end-to-end manner.

Despite the significant progress achieved by CNN- and Transformer-based approaches in fingerprint embedding, we recognize that \textit{previous works do not exploit nonstructural information, such as topology and correlation in fingerprints}, which is significant to facilitate the identifiability and robustness of AFIS. It is worth noting that fingerprint data possess rich non-grid structures, such as the distances and adjacency relationships between the minutiae scattered on fingerprints. Although CNN and Transformer excel at learning salient features from the grid-like pixel data in images, it is hard for them to exploit the non-grid structure of fingerprints and mine the valid topological information due to their inherent limitations.  In this paper, we present a novel paradigm for fingerprint embedding, called \textit{Minutiae Relation-Aware model over Graph Neural Network} (MRA-GNN), which incorporates a GNN-based framework in fingerprint embedding to encode the topology and correlation of fingerprints into descriptive features.

In terms of technique, MRA-GNN is equipped with \textit{Topological relation Reasoning Module} (TRM) and \textit{Correlation-Aware Module} (CAM) to learn the fingerprint embedding from both minutia-level and fingerprint-level graphs rather than image pixels.  As shown in Fig. \ref{fig:graph construct}(a), \textit{we reinterpret minutiae and their connections as vertices and edges} respectively, and a graph structure is introduced to represent the topology of each fingerprint. Then the graph is utilized to infer implicit topological relations among minutiae by means of the GNN-based TRM. In Fig. \ref{fig:graph construct}(b), \textit{a batch of fingerprints is considered as vertices, and the similarities of their global features are adopted to construct edges}, resulting in the formation of a graph,  which is then applied to perceive inherent correlation structures among fingerprints via the GNN-based CAM. Benefiting from the GNN-based framework, our proposed approach can be efficiently and accurately trained in an end-to-end way with the supervision of triplet loss. To alleviate the over-smoothing problem of GNN, we first incorporate graph residual connections into GNN-based TRM and CAM, then propose to apply Feed-Forward Module (FFM) after them to maintain the diversity of vertex features, enabling the greater performance of graph embedding as layers deepen.  
 
To our best knowledge, this work is the first to successfully apply graph neural network to achieve fingerprint embedding according to topological information and correlation structures. We believe this novel paradigm will inspire the community to further explore GNN-based models in fingerprint embedding. In a nutshell, the main contributions of this paper are: 
\begin{itemize}
    \item We propose a novel graph-based paradigm for fingerprint embedding. The final representation is extracted from the topological graphs of minutiae and fingerprints rather than image pixels employed in previous deep learning works. 
    \item We devise an MRA-GNN model under the paradigm, which is the first to study compact and identifiable fingerprint embedding via a deep GNN for large-scale AFIS. Since topological relations among minutiae and correlation structures among fingerprints are considered, it is intuitive that the embedding of MRA-GNN has stronger robustness and generalization.  
    \item Extensive experiments demonstrate that the proposed MRA-GNN is eminently competitive with prior state-of-the-art methods \cite{Fingerpatches,Deepprint,LatentAFIS,transformer-based2}. For instance, MRA-GNN achieves the best TAR@FAR=0.1\% on NIST SD4 (99.15\%) and EER on FVC2004 DB1 (0.53\%) in recognition task, and performs best on NIST SD4 dataset in indexing task.
 \end{itemize} 


\begin{figure*}
	\begin{center}
		\includegraphics[width=1\linewidth]{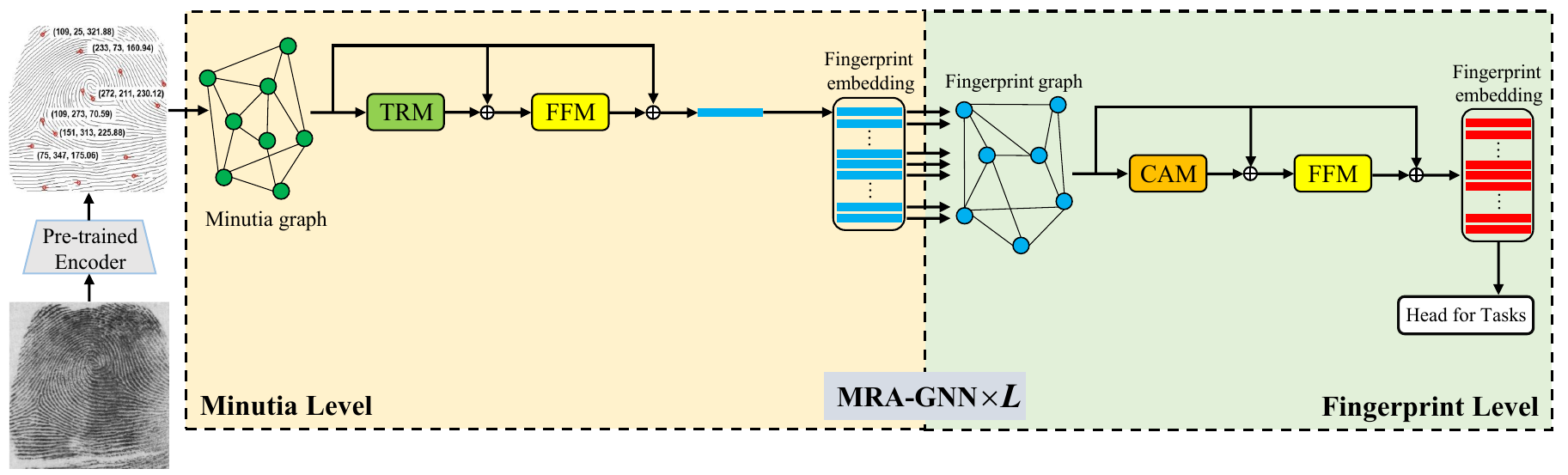}
	\end{center}
	\caption{Illustration of the proposed MRA-GNN for fingerprint embedding. The framework includes: (1) Minutia level performs the minutia-based graph construction and the reasoning of implicit topological relations. (2) Fingerprint level achieves the fingerprint-based graph construction and the awareness of inherent correlation structures. The embedding can be applied to various fingerprint tasks.}
	\label{fig: pipeline}
\vspace{-1em}
\end{figure*}

\vspace{-0.8em}
\section{Related work}
\paragraph{Fingerprint embedding approach.}
Fingerprint feature embedding plays a critical role in modern AFIS. Classical embedding methods are generally based on minutia triplets \cite{triplets-1,triplets-2}or cylinder-code \cite{cylinder-1}. However, algorithms with these geometric structures heavily rely on handcrafted rules and complex design procedures that depress the performance of representation. With the emergence of deep learning, the research on deep feature embedding is increasing \cite{Fingerpatches, LatentAFIS, Deepprint, MaRs}. These methods do not need to undergo cumbersome manual design as classical ones. More precisely, Li \textit{et al}. \cite{Fingerpatches} adopt paired fingerprint patches centered on minutiae to extract local descriptors which will be aggregated into a fixed-length representation through global average pooling. Nevertheless, the method suffers from a lack of generalization and interpretability, and minutia information is not well learned by simple global pooling. Follow-up method \cite{LatentAFIS} presents an end-to-end latent fingerprint search system with two separate minutia extraction models to provide complementary minutia templates and texture templates. Besides, Engelsma \textit{et al}. \cite{Deepprint} subsequently learn to extract fixed-length fingerprint embedding based on global texture information with a multi-task CNN. Their representations mainly focus on global or local patterns of fingerprints but do not pay attention to the topology and correlation in them. Thus, the above algorithms lack compact and discriminative features. A new approach to fingerprint embedding needs to emerge urgently. \vspace{-1em}

\paragraph{Graph neural network.}
Our work is also related to graph neural network (GNN), which has been successfully applied to various scenarios in deep learning \cite{GNN-application1}. The earliest GNN was initially outlined in \cite{initial-GNN1}. Micheli \cite{spatial-GNN} proposes the form of spatial-based GNN by architecturally compositing non-recursive layers. Spectral-based graph convolutional network was first presented by Bruna \textit{et al}. \cite{spectral-GNN} who introduce graph convolution based on the spectral graph theory. GCN is usually employed in graph data, such as social networks and biochemical graphs \cite{GNN-socialnetwork}. In recent years, it is gradually applied to computer vision, natural language processing, and other fields \cite{ViG, LSCM-GNN}. Nevertheless, these methods are at the primary stage. They are prone to over-smoothing, have inferior accuracy and robustness, and the interface with tasks is not perfect. Different from existing works on GNN, which mainly focus on graph-structure data or scene objects, we introduce it to characterize topological relations and correlation structures in fingerprints. As far as we know, our work is the first to study compact and discriminative fingerprint embedding on GNN.


\section{Methodology}

In this section, we introduce the graph-based paradigm that incorporates a GNN-based framework within fingerprint embedding to encode the topology and correlation of fingerprints into descriptive features. In the following subsections, we first illustrate the basic knowledge of GNN (Sec. \ref{sec:pre}). Then, we introduce the design philosophy of MRA-GNN (Sec. \ref{sec:overall}), in which Topological relation Reasoning Module (Sec. \ref{sec:trm}) and Correlation-Aware Module (Sec. \ref{sec:cam}) learn the fingerprint embedding from both minutia-level and fingerprint-level graphs rather than texture patterns or image pixels. Lastly, the Feed-Forward Module and graph residual connections (Sec. \ref{sec:oversmooth}) for tackling the over-smoothing problem of GNN are presented.

\subsection{Preliminaries}\label{sec:pre}

\paragraph{Graph Convolutional Network.} Given a graph $\mathcal{G}=(\mathcal{V},\mathcal{E})$ with $N$ nodes $v_i \in \mathcal{V}$ and edges $(v_i, v_j)\in \mathcal{E}$, the Graph Convolutional Network (GCN), which is composed of graph convolutional layers, is designed to obtain an embedding of each vertex by fusing information from the graph. Specifically, let $x_i$ be the feature vector of $v_i$, and $X = [x_1, \dots, x_N]$, then a graph convolutional layer can be defined as
\begin{equation}\label{eq: gcn}
\mathrm{GraphConv}(X;\mathcal{G})= \sigma(h \circ g(X)),
\end{equation}
where $g$ is an \textit{aggregation} operator to aggregate features of neighbors within $\mathcal{G}$, $h$ is an \textit{update} function with learnable parameters to map aggregated features into another space, and $\sigma$ is an activation to increase the non-linearity of features. For example, Kipf \textit{et al}. \cite{Kipf2016SemiSupervisedCW} propose a simplified and efficient GCN layer using
$g: X \rightarrow \hat{A}X, h: X \rightarrow XW, \sigma: X \rightarrow ReLU(X)$,
where $\hat{A}$ is a normalized adjacency matrix derived from $\mathcal{E}$. With these definitions, the GCN layer can be written as
\begin{equation}
\mathrm{GraphConv}(X;\mathcal{G})= \mathrm{ReLU}(\hat{A}XW).
\end{equation}
For more details, we refer readers to the seminal work \cite{Kipf2016SemiSupervisedCW}.\vspace{-1em}

\paragraph{Over-smoothing phenomenon} is a prevalent issue in various GCNs~\cite{over-smoothing}. As graph convolutional layers deepen, the feature vectors of different nodes tend to converge towards similar representation, resulting in performance degradation of the GCN model. This issue restricts the capacity of GCN for modeling complex graph structures, and applications of GCN for large-scale datasets.

In our work, we propose to use graph to express the topology of fingerprints and introduce a GCN-based model to extract informative embedding from fingerprints. 

\subsection{Overview of MRA-GNN}\label{sec:overall}
Fig. \ref{fig: pipeline} provides an overview of the proposed MRA-GNN for graph-based fingerprint embedding, which contains minutia level and fingerprint level. Each level possesses two steps: graph construction and graph embedding. Specifically, (\textbf{I}) at the minutia level, we first adopt an off-the-shelf backbone $\boldsymbol{E}$ to extract minutiae from fingerprint $F$. Subsequently, we regard each minutia $\boldsymbol{E}(F)$ as a central vertex and find its $\mathcal{K}_{m}$ nearest neighbors to construct a minutia graph $\mathcal{G}_{m}(F)$. Then we utilize TRM to infer  minutia graph embedding $m_{F}$ with topological relations implied in the graph. We abbreviate this process as
\begin{equation}\label{eq1}
m_{F} = TRM(\mathcal{G}_{m}(F)).
\end{equation}
(\textbf{II}) At the fingerprint level, we sample a batch of fingerprints and find $\mathcal{K}_{f}$ nearest neighbors of fingerprint $F$ based on minutia graph embedding $m_{F}$ to construct graph $\mathcal{G}_{f}(F)$. Then we adopt CAM to compute fingerprint graph embedding $M_{F}$ with correlation structures as fingerprint representation. We abbreviate this process as
\begin{equation}\label{eq2}
M_{F}=CAM(\mathcal{G}_{f}(F)).
\end{equation}
(\textbf{III}) Since the increase of GCN layers in TRM and CAM will inevitably cause the over-smoothing problem, we introduce FFM and graph residual connections to the design of MRA-GNN, in order to enhance the model capacity while preventing features from collapsing. 

In summary, for a given fingerprint, MRA-GNN extracts its embedding from a couple of minutia graphs and fingerprint graphs by means of anti-smoothing TRM and CAM.

\subsection{Topology reasoning on minutia graph}\label{sec:trm}

In this section, we detail the minutia level based on Topological relation Reasoning Module (TRM), which works on minutia graph construction and implicit topological relation inference among minutiae. Thanks to the efficiency and robustness, minutia-based features that fully represent fingerprints, are extensively applied in fingerprint embedding rather than texture features and image pixels. Previous works are generally based on minutia geometries and image patches and do not exploit topological relations among minutiae. The advantages of minutia topology based on graph paradigm include: 1) graph is a generalized data structure that fixed geometry and grid can be regarded as a special case of graph; 2) graph is more flexible and precise than fixed geometry or gird to model irregular structures among minutiae in a fingerprint; 3) similar information can be aggregated and updated by the constructed graph connections among minutiae, thus enhancing the performance of minutia embedding; 4) the advanced research on GNN can be transferred to address fingerprint tasks.\vspace{-1em}

\paragraph{Minutia graph construction.} For each fingerprint $F$, we first employ an off-the-shelf minutia extractor $\boldsymbol{E}$~\cite{minutiae_extractor} to extract minutiae in $F$ and transform each of them into a 3-dimension feature vector $x_{i}\in\mathbb{R}^{3}$ that consists of location and orientation information $(x, y, d)$, \textit{i.e.}, $\boldsymbol{E}(F)=[x_{1},x_{2},\cdots,x_{N}]$. These minutiae can be viewed as a set of unordered vertices, which are denoted as $\mathcal{V}_{m}=\{v_{1},v_{2},\cdots,v_{N}\}$. For each vertex $v_{i}$, we find its $\mathcal{K}_{m}$ nearest neighbors $\mathcal{N}(v_{i})$ and add an unoriented edge $(v_i, v_j)$ to $\mathcal{E}_{m}$ for every $v_j\in \mathcal{N}(v_{i})$. The visualized construction process of the minutia graph is illustrated in Fig. \ref{fig:graph construct}(a). After constructing minutia graph structure $\mathcal{G}_{m}(F)=(\mathcal{V}_{m},\mathcal{E}_{m})$ of $F$, we explore how to adopt GNN to deduce implicit topological relations among minutiae.\vspace{-1em}

\begin{figure}[t]
	\begin{center}
		\includegraphics[width=1\linewidth]{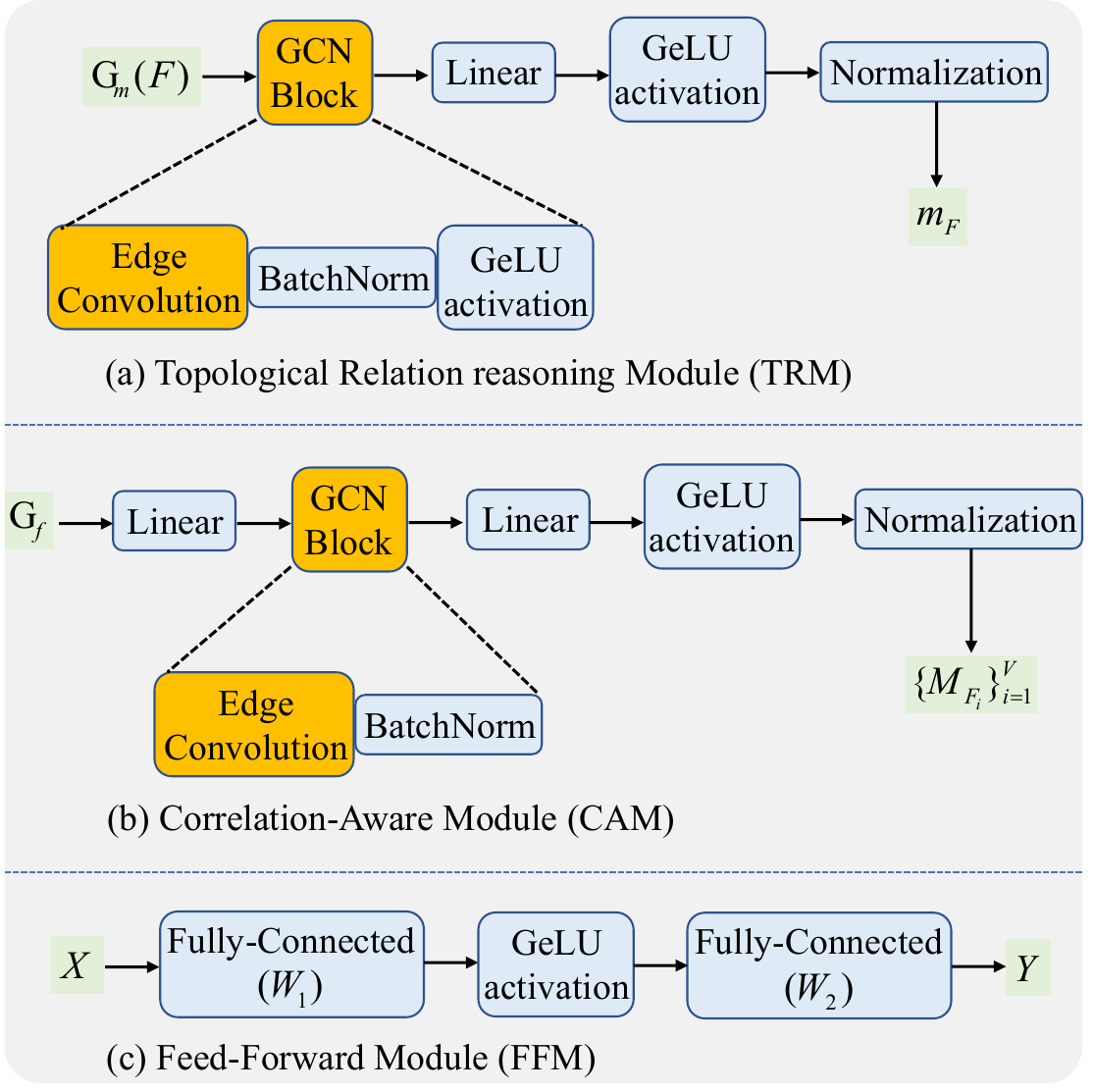}
	\end{center}
	\caption{Illustration of various modules in MRA-GNN.
 }
    \vspace{-1em}
	\label{fig: module}
\end{figure}

\paragraph{TRM-based minutia graph embedding.}  As shown in Fig. \ref{fig: module}(a), the input minutia graph $\mathcal{G}_{m}(F)$ of TRM is composed of minutia relations and their features. TRM consists of one GCN block, one linear layer with GeLU activation\cite{GeLU}, and one normalization layer. To fully reason the topological relations among vertices, the GCN block is basically made up of EdgeConv graph convolutional layer\cite{Edgeconv}, followed by a batch normalization layer and GeLU activation. Here EdgeConv is employed for performing a channel-wise symmetric aggregation operation on both vertex features and edge features associated with all edges originating from each vertex. As introduced in Eq. \eqref{eq: gcn}, the aggregation function of EdgeConv is defined as 
\begin{equation}\label{eq: edgeconv}
	g(x_{i})=\mathrm{\mathop{concat}}(x_{i}, \max(\{e_{ij}|(v_i,v_j)\in \mathcal{E}\})).
\end{equation}
Here, $\mathrm{\mathop{concat}}$ denotes the concatenation of input vectors, and the edge feature $e_{ij}$ is given by the equation:
\begin{equation}
	e_{ij}={\rm GeLU}(\theta \cdot (x_{j}-x_{i})+\phi \cdot x_{i}),
\end{equation}
where $\theta$ and $\phi$ are learnable parameters. The update function is a linear transformation $h(x) = xW$, as used in previous works. After the GCN block, we employ a linear layer to unify the dimensions of graph embedding, so that each fingerprint has a fixed-length feature. We integrate all vertex features from TRM to be the embedding $m_{F}$ in Eq. \eqref{eq1}.

\subsection{Correlation learning on fingerprint graph}\label{sec:cam}

In this section, we detail the Correlation-Aware Module (CAM), which works on fingerprint graph construction and inherent correlation structure awareness among fingerprints. Based on the graph paradigm, similar fingerprints can be aggregated and updated by correlation information, while different fingerprints are structurally distanced, thus improving the performance of feature embedding.\vspace{-1em}

\paragraph{Fingerprint graph construction.} Considering a batch of fingerprints $\{F_{i}\}_{i=1}^{V}$ and their features $\{m_{F_{i}}\}_{i=1}^{V}$, we view them as vertices to express the local structure in fingerprint manifold and adopt minutia graph embeddings as the initial vertex features. To obtain correlation information in the structure, for each fingerprint $F_{i}$ we utilize K-NN algorithm to add unoriented edges $e_{ij}$ for all $F_{j}\in \mathcal{N}(F_{i})$ by Euclidean distance. After that, we obtain a fingerprint graph $\mathcal{G}_{f}$. The visualized construction process of the fingerprint graph is shown in Fig. \ref{fig:graph construct}(b). In the following words, we explore how GNN can be applied to perceive correlation structures among fingerprints.\vspace{-1em}

\paragraph{CAM-based fingerprint graph embedding.} Fig. \ref{fig: module}(b) displays the details of Correlation-Aware Module (CAM). The input of CAM is graph $\mathcal{G}_{f}$ that is composed of a batch of fingerprints and their initial features. CAM consists of one GCN block, two linear layers, and one GeLU activation layer with normalization. Here we employ the same EdgeConv as Eq. \eqref{eq: edgeconv} in the GCN block, followed by a batch normalization layer. 
The output of the entire fingerprint level is the graph embedding features $\{M_{F_{i}}\}_{i=1}^{V}$, in which $M_{F_{i}}$ is the corresponding feature embedding of fingerprint $F_{i}$ as denoted in Eq. \eqref{eq2}.


\subsection{Over-smoothing alleviation}\label{sec:oversmooth}

It is difficult to scale up GCNs due to the over-smoothing problem, thus we introduce two additional ways to mitigate this problem in the MRA-GNN.
\vspace{-1em}

\paragraph{Feed-Forward Module.} First, we insert Feed-Forward Module (FFM) into MRA-GNN to alleviate over-smoothing problem. FFM is a multi-layer perceptron with two Fully-Connected (FC) layers and one GeLU activation layer, as illustrated in Fig. \ref{fig: module}(c). Concretely, FFM is given by
\begin{equation}
	Y = {\rm GeLU}(XW_{1})W_{2},
\end{equation}
where input $X$ can be $m_{F}$ and $\{M_{F_{i}}\}_{i=1}^{V}$ at minutia and fingerprint levels, respectively. $W_{1}$ and $W_{2}$ are the weight of FC layers. Through the FC layers, we can project node features into the same domain and increase the feature diversity, thus the embedded vectors of different nodes will converge to different representations. Moreover, layer collapse will be avoided by nonlinear activation.\vspace{-1em}

\paragraph{Graph residual connection.}Inspired by the huge success of ResNet \cite{ResNet}, we transfer residual connections to GCN, thus unleashing its potential. In the original GCN framework as Eq. \eqref{eq: gcn}, the underlying mapping $\sigma(h \circ g)$, which takes a graph as input and outputs a new graph representation, is learned. Here we introduce graph residual connection in the original GCN framework. Specifically, after $X$ is transformed by $\sigma(h \circ g)$, vertex-wise addition is performed to obtain the graph residual connection, which is defined as 
\begin{equation}\label{eq: res_gcn}
\mathrm{GraphConv}_{res}(X;\mathcal{G})= \sigma(h \circ g(X)) + X.
\end{equation}
The graph residual connection can reduce the back-propagation complexity of GCN and prevent gradient vanishing, which enables deeper GCN to reliably converge in training and achieve superior performance in inference.

\begin{figure}[t]
	\begin{center}
		\includegraphics[width=1\linewidth]{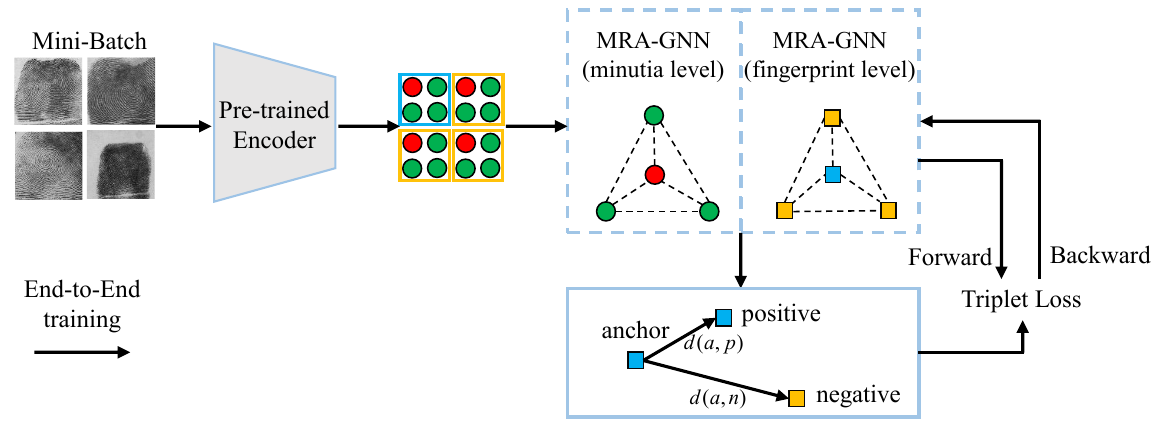}
	\end{center}
	\caption{The detail training process of our framework. The backbone encoder is first pre-trained, then we train the whole network end-to-end with the supervision of triplet loss.}
	\label{fig: training}
\vspace{-1em}
\end{figure}

\begin{table*}
	\begin{center}
		\begin{tabular}{*{5}{c}}
			\toprule
			\multirow{2.5}*{Methods} & \multicolumn{3}{c}{TAR@FAR=0.1\%} & \multicolumn{1}{c}{EER}\\
			\cmidrule(lr){2-4}\cmidrule(lr){5-5}
			& FVC2004 DB1 & NIST SD4 & NIST SD14 & FVC2004 DB1 \\
			\midrule
			FingerPatches\cite{Fingerpatches} & - & 96.57\% & 97.30\% & - \\
			DeepPrint\cite{Deepprint} & 97.50\% & 97.90\% & 98.55\% & 1.75\%\\
			LatentAFIS\cite{LatentAFIS} & 97.00\% & 98.70\% & 99.00\% & 1.31\%\\
			TMNet\cite{TMNet} & 98.04\% & 97.46\% & 98.24\% & 1.02\% \\
			Convolution-Transformer\cite{transformer-based2} & \textbf{98.55\%} & 98.35\% & 99.13\% & 0.84\%\\
			MRA-GNN(ours) & 98.35\% & \textbf{99.15\%} & \textbf{99.60\%} & \textbf{0.53\%}\\
			\bottomrule
		\end{tabular}
	\end{center}
	\caption{The comparisons of TAR@FAR=0.1\% and EER in existing methods and the proposed model for recognition tasks on three benchmarks: NIST SD4, NIST SD14, and FVC2004 DB1. '-' denotes that previous methods do not present this indicator and their codes are not released, with difficulties of reproducing based on the description of the paper. The best results are highlighted in bold.}
 \vspace{-1.2em}
        \label{table: recognition}
\end{table*}


\subsection{Training}
The detailed training scheme is shown in Fig. \ref{fig: training}, in which we optimize parameters of MRA-GNN under the supervision of triplet loss\cite{tripletloss} in an end-to-end manner. Considering a batch of fingerprints, we first select the most similar fingerprint pairs as the anchor-positive pairs $(a,p)$. Next, for each pair $(a,p)$ we choose the hardest sample $n$, which is most similar to the anchor but has a different label, to generate semi-hard triplet $(a,p,n)$. The target of triplet loss is minimizing the distance of anchor-positive pairs $d(a,p)$ while maximizing the distance of anchor-negative pairs $d(a,n)$. Here we adopt $l_{2}-norm$ of feature embedding to measure the distance between fingerprint pairs$(f_{1}, f_{2})$, which is known as $d(f_{1},f_{2})=||f_{1}-f_{2}||_{2}$. In summary, the objective function of MRA-GNN training is given by
\begin{equation}\label{eq: trip loss}
	Loss(a,p,n) = \max(d(a,p)-d(a,n)+\rm{\gamma}, 0),
\end{equation}
where the constant $\gamma$ is a margin that forces the model to learn smaller $d(a,p)$ and larger $d(a,n)$. The loss ensures that the model can learn diverse features from samples.

\section{Experiments}

In this section, we report the datasets and different experiments that we performed for testing the MRA-GNN model accompanied by the results. Since our proposed method achieves a novel graph paradigm for fingerprint embedding, which is a framework capable of being generalized to any end-to-end fingerprint task, we conduct experiments in fingerprint recognition and indexing tasks, respectively.

\subsection{Datasets}

For the training and validation, we adopt an in-house dataset comprised of 109K rolled fingerprint images captured by optical sensors from 11,265 unique fingers with the spatial size $256\times360$ , in which 90K and 19K are utilized for each target. For testing, the widely used benchmark NIST SD4\cite{nist-sd4} and NIST SD14\cite{nist-sd14} that are made up of rolled fingerprint images similar to the training dataset, as well as slap fingerprint datasets FVC2004 DB1\cite{fvc2004}, are applied in the recognition tasks. In terms of indexing tasks, since the top-5 accuracy of NIST SD14 has been elevated to 100\% by MaRs\cite{MaRs} and without the necessity to optimize, we adopt NIST SD27\cite{nist-sd27} that contains 258 pairs of latent and rolled fingerprints instead. There are 2K rolled fingerprint pairs in NIST SD4 with the spatial size $512\times512$, and 27K rolled fingerprint pairs in NIST SD14 with the spatial size $832\times768$. However, only the last 2,700 pairs from NIST SD14 are used for evaluation in previous approaches, and we employ them for fair comparison. The major motivation for selecting FVC2004 DB1 is that it is comprised of slap fingerprint images and is extremely challenging. Hence, we are able to demonstrate that even if MRA-GNN is trained on rolled fingerprint images, our incorporation of domain knowledge into the framework enables itself to be robust and generalize well to slap fingerprint datasets.


\subsection{Implementation details}

We adopt dilated aggregation\cite{DeepGCNs} in both TRM and CAM modules, and set the dilated rate as $\lceil l/4 \rceil$ for the $l$-th layer. To fair comparison, we apply a batch size of 256 to train our model on GeForce RTX 3090 for 200 epochs with AdamW optimizer. The learning rate is 1e-4 in the initial stage and varies according to the cosine schedule. The default values of momentum and weight decay factor of AdamW are employed. The margin $\gamma$ is set at $0.5$. During training, we augment the dataset with random rotations (from$-15^{\circ}$ to $15^{\circ}$) and translations. As for fingerprint alignment, the topological invariance of graph allows our method to mitigate the effects of geometric transformation.


\subsection{Fingerprint recognition}
In the fingerprint recognition stage, each fingerprint is first represented as an embedded vector by MRA-GNN. Subsequently, the similarity of two fingerprints can be measured through the product of the corresponding embedding, which is defined as
\begin{equation}
	Sim(F_{i}, F_{j})=<M_{F_{i}}, M_{F_{j}}>,
\end{equation}
where $F_{i}$ and $F_{j}$ denote two fingerprints, $M_{F_{i}}$ and $M_{F_{j}}$ indicate the fingerprint feature embedding extracted by MRA-GNN. Thanks to the normalization of representation in CAM, the product of two vectors is similar to cosine similarity. For a given fingerprint, we can determine the match by setting a fixed similarity threshold, thereby identifying whether fingerprint pairs come from the same finger or not.

For evaluating the recognition performance of our MRA-GNN fingerprint embedding model, we make a fair comparison with previous approaches on FVC2004 DB1, NIST SD4, and NIST SD14. In terms of evaluation metrics, we adopt TAR@FAR=0.1\% for each benchmark, which indicates the True Accept Rate (TAR) when False Accept Rate (FAR) is 0.1\%. Apart from that, Equal Error Rate (EER) which is the error rate when FAR and False Reject Rate (FRR) are equal, is utilized to evaluate the performance of FVC2004 DB1 given its specific properties.

The comparisons of experimental results of fingerprint recognition are presented in Table \ref{table: recognition}. Compared with five other algorithms such as DeepPrint\cite{Deepprint} and LatentAFIS\cite{LatentAFIS}, the performance of our proposed MRA-GNN surpasses most of them. Especially when measured in the TAR@FAR=0.1\% indicator, our proposed method achieves 99.15\% and 99.60\% on NIST SD4 and NIST SD14 respectively, which is significantly better than all other methods. For metrics on FVC2004 DB1, we notice that the TAR@FAR=0.1\% of our method has a small drop of 0.2\%. However, the EER metric outperforms SOTA by 0.31\%, which demonstrates that MRA-GNN has the ability to be robust and generalize well over various fingerprint datasets.

As described above, the performance improvement of MRA-GNN on fingerprint recognition is prominent, especially on NIST datasets. In the following section, we perform a series of ablation studies to analyze the influence of graph convolutional types, components, and different configurations of our proposed MRA-GNN framework.


\subsection{Ablation studies}

We conduct ablation studies of the proposed model on fingerprint recognition tasks of NIST SD4 and NIST SD14.\vspace{-1em}

\begin{table}
	\begin{center}
		\begin{tabular}{*{2}{c}}
			\toprule
			GraphConv type & TAR@FAR=0.1\% \\
			\midrule
			GraphSAGE\cite{GraphSAGE} & 98.15\% \\
			  GIN\cite{GIN} & 98.43\%\\
			Max-Relative GraphConv\cite{DeepGCNs} & 98.76\% \\
			EdgeConv\cite{Edgeconv} & \textbf{99.15\%} \\
			\bottomrule
		\end{tabular}
	\end{center}
	\caption{Ablation studies on different types of graph convolution. }
        \label{table: ablation graphconv}
\end{table}

\begin{table}
	\begin{center}
		\begin{tabular}{*{4}{c}}
			\toprule
			TRM & CAM & FFM & TAR@FAR=0.1\% \\
			\midrule
			\checkmark &   &   & 95.83\% \\
			 & \checkmark &  & 96.17\%\\
			\checkmark & \checkmark &  & 98.32\% \\
			\checkmark &  & \checkmark & 97.15\% \\
			 & \checkmark & \checkmark & 97.68\% \\
			\checkmark & \checkmark & \checkmark & \textbf{99.15\%} \\
			\bottomrule
		\end{tabular}
	\end{center}
	\caption{Ablation studies on different modules of MRA-GNN.}
        \label{table: ablation module}
\end{table}

\begin{table}
	\begin{center}
		\begin{tabular}{*{3}{c}}
			\toprule
			\multirow{2.5}*{Batch size} &             
                \multicolumn{2}{c}{TAR@FAR=0.1\%}\\
			\cmidrule(lr){2-3}
			& NIST SD4 & NIST SD14\\
			\midrule
			64 & 99.12\% & 99.45\%\\
			128 & 99.09\% & 99.52\% \\
			256 & \textbf{99.15\%} & \textbf{99.60\%} \\
                320 & 99.06\% & 99.42\% \\
			512 & 98.97\% & 99.48\% \\
			\bottomrule
		\end{tabular}
	\end{center}
	\caption{Ablation studies on different training batch sizes of MRA-GNN. We record the TAR@FAR=0.1\% on NIST SD4 and NIST SD14 to verify the performance of our framework.}
\vspace{-1.5em}
        \label{table: ablation batchsize}
\end{table}

\paragraph{The effects of graph convolutional types.}We test the representative variants of graph convolution for TRM and CAM, including GraphSAGE\cite{GraphSAGE}, GIN\cite{GIN}, Max-Relative GraphConv\cite{DeepGCNs} and EdgeConv\cite{Edgeconv}. From Table \ref{table: ablation graphconv}, we conclude that the TAR@FAR=0.1\% accuracies of different graph convolutions on NIST SD4 are competitive with SOTA, indicating the flexibility of MRA-GNN architecture. Among them, EdgeConv achieves the best performance. In the following experiments, we adopt EdgeConv as the graph convolutional layer by default unless specially stated.\vspace{-1em}

\begin{table*}
	\begin{center}
		\begin{tabular}{*{7}{c}}
			\toprule
			\multirow{2.5}*{Methods} & \multicolumn{3}{c}{NIST SD4} & \multicolumn{3}{c}{NIST SD27}\\
			\cmidrule(lr){2-4}\cmidrule(lr){5-7}
			& Top-1 & Top-5 & Top-10 & Top-1 & Top-5 & Top-10 \\
			\midrule
			FingerPatches\cite{Fingerpatches} & 99.27\% & \textbf{99.65\%} & \textbf{99.77\%} & - & - & - \\
			DeepPrint\cite{Deepprint} & 98.70\% & 99.22\% & 99.75\% & 65.50\% & 69.84\% & 72.15\% \\
			LatentAFIS\cite{LatentAFIS} & 98.55\% & 98.94\% & 99.15\% & 68.60\% & 73.30\% & 75.10\%\\
			Multi-scale representation\cite{Multi-scale} & 98.80\% & - & - & 69.38\% & - & - \\
			MaRs\cite{MaRs} & 99.35\% & \textbf{99.65\%} & 99.70\% & - & - & -\\
			MRA-GNN(ours) & \textbf{99.46\%} & \textbf{99.65\%} & 99.75\% & \textbf{70.50\%} & \textbf{74.45\%} & \textbf{76.84\%}\\
			\bottomrule
		\end{tabular}
	\end{center}
	\caption{Top-k accuracy for fingerprint indexing on two benchmarks: NIST SD4 and NIST SD27. '-' denotes that previous methods do not present this indicator and their official codes are not released. The best results are highlighted in bold.}
        \label{table: indexing}
\vspace{-1em}
\end{table*}

\paragraph{The effects of modules in MRA-GNN.} To make GNN adaptive to the proposed minutia-based fingerprint embedding model, we have introduced three modules: TRM, CAM, and FFM in the above sections. TRM and CAM are adopted to mine deep topological structures in fingerprints, and FFM is employed in transforming features and alleviating the over-smoothing of GNN. All three constitute MRA-GNN architecture. To fully evaluate the effects of these modules on fingerprint embedding performance, ablation studies are conducted for them on NIST SD4. We change the feature dimensions of the compared models to make their FLOPs similar, so as to have fair comparison. Table \ref{table: ablation module} demonstrates that using only one of the two GCN modules for fingerprint recognition performs poorly while combining these two delivers significant performance gains. For the effect of FFM, we discover that, to mitigate the over-smoothing problem, introducing FC layers and graph residual connections consistently increases the TAR@FAR=0.1\% metric. In summary, the integration of the above three modules can maximize the advantages of fingerprint embedding based on graph paradigm.\vspace{-1.2em}

\paragraph{The effects of batch size.} Graph construction at fingerprint level allows CAM to extract fingerprint embeddings in different manifolds. Batch size controls the diversity of information transformation in manifolds and the scale of fingerprint graph. The experimental results for different batch sizes are displayed in Table \ref{table: ablation batchsize}. TAR@FAR=0.1\% indicator on both datasets changes slightly with different batch sizes, which demonstrates the robustness and stability of our proposed MRA-GNN. We select 256 as the default value for optimal trade-off between accuracy and efficiency.\vspace{-1em}

\begin{figure} [t]
	\centering
	\includegraphics[width=1\linewidth]{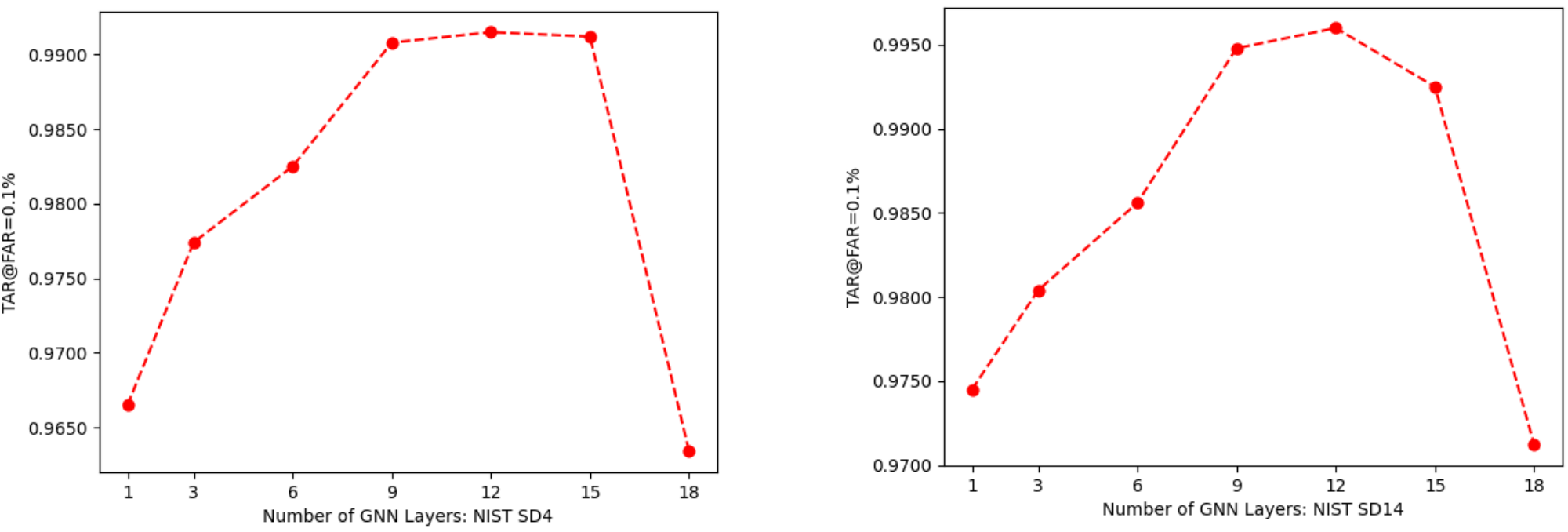}
	\caption{The effects of different GNN layers.}
	\label{fig: layers}
\end{figure}

\paragraph{The effects of the number of GNN layers.} In the process of constructing GNN, the number of graph convolutional layers $L$ is a vital hyperparameter, which impacts the feature embedding performance of the architecture. Too few layers will degrade the capability of embedding, while too many layers will lead to over-smoothing that is unable to avoid by FFM. We tune $L$ from 1 to 18 and conduct experiments on NIST SD4 and NIST SD14, respectively. The results are presented in Fig. \ref{fig: layers}. We discover that the indicator tends to increase and then decrease as $L$ rises, and performs best when $L$ is in the range of 9 to 15.

\begin{figure} [t]
	\centering
	\includegraphics[width=1\linewidth]{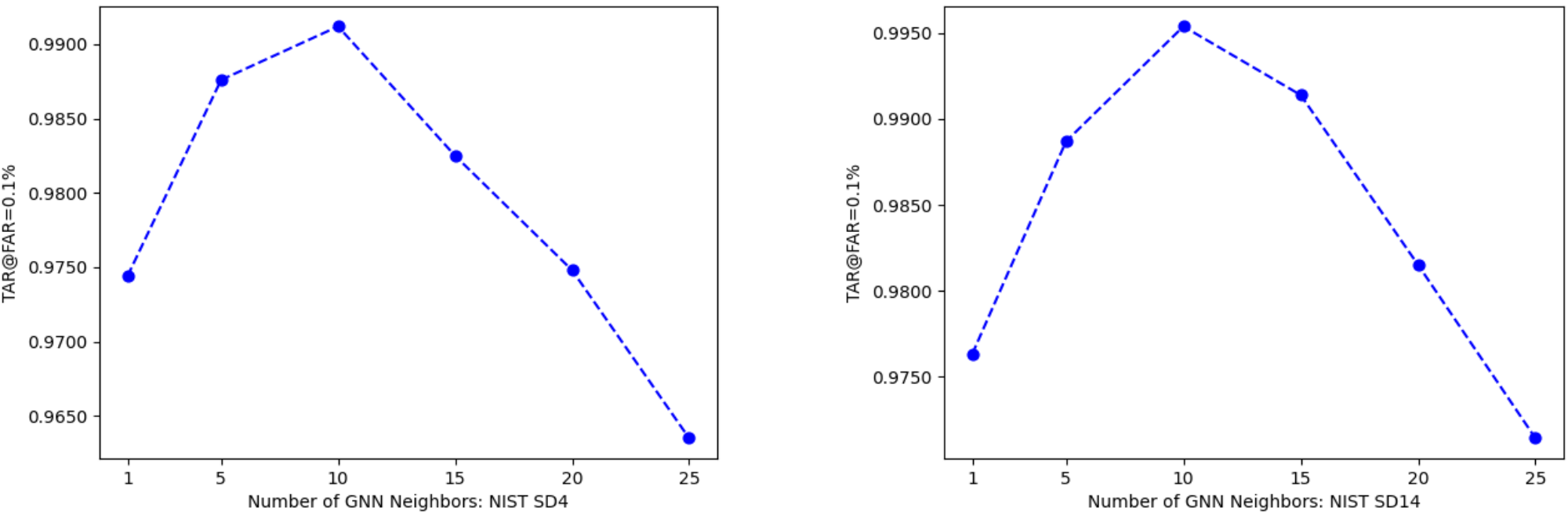}
	\caption{The effects of different GNN neighbors.}
	\label{fig: neighbors}
 \vspace{-1em}
\end{figure}

\paragraph{Impact of the number of GNN neighbors.} The performance of our model is also related to the number of neighbor nodes $K$ in graph, which controls the aggregated range. Too few neighbors will restrain information exchange, while too many neighbors will also lead to inevitable over-smoothing and more irrelevant vertices will be included. Fig. \ref{fig: neighbors} displays the TAR@FAR=0.1\% obtained by MRA-GNN with different values of $K$ on NIST SD4 and NIST SD14. We discover that the performance of our proposed method is increasing with $K$ in the initial stage, and reaches the maximum at around $K=10$.

\subsection{Fingerprint indexing}

We employ MRA-GNN for fingerprint indexing, which selects candidate sets for a given fingerprint based on the similarity of feature embeddings. Particularly, for a query fingerprint, we filter out the candidate set of a certain length from the gallery according to embedded vectors, hence rapidly identifying whether fingerprint pairs come from the same finger or not and reducing the search space for indexing. The gallery we adopted is a closed set (each probe has a mate in gallery), which contains 10K rolled fingerprints. For a fair comparison, we report the Top-k accuracy of indexing on NIST SD4 and NIST SD27, which is more precise and stricter than the traditional error rate for a given penetration rate. From the results in Table \ref{table: indexing}, we conclude that our method performs best on the NIST SD27 given that MRA-GNN only requires identifiable sub-graph structures of the fingerprints. Moreover, the Top-1, Top-5, and Top-10 accuracy on NIST SD4 are able to reach 99.46\%, 99.65\%, and 99.75\%, which maintain around the optimal indicators.

\section{Conclusion}

In this paper, we pioneer studying the topology of fingerprints by deep graph-paradigm embedding. We develop an MRA-GNN model to extract fingerprint embedding with superior performance. At the minutia level, we reinterpret minutiae and their connections as vertices and edges for minutia graph construction, and introduce TRM to reason implicit topological relations among minutiae. At the fingerprint level, a batch of fingerprints is considered as vertices and we design CAM to perceive inherent correlation structures among fingerprints. For alleviating the over-smoothing problem, we propose FFM and graph residual connections to maintain feature diversity as layers deepen. Extensive experiments demonstrate the superiority, robustness, and generalization of our framework. 

Admittedly, the main limitation of our study is the demand for high-quality initial features, and our model currently performs on fingerprint datasets but cannot execute online matching for a single fingerprint. For future work, exploring more effective graph models and applying them to broader biometrics will be meaningful.
\paragraph{Acknowledgements}
This work is supported by the National Natural Science Foundation of China (12271504).

{\small
\bibliographystyle{ieee}
\bibliography{egbib}
}
\end{document}